\documentclass[letterpaper, 10 pt, conference]{ieeeconf}  

\IEEEoverridecommandlockouts                              

\usepackage{amsmath,graphicx}
\usepackage{mathrsfs}
\usepackage{multirow}  
\usepackage{cite}
\usepackage{amsfonts,amssymb}
\usepackage{stfloats}
\usepackage{float}
\usepackage{booktabs}
\usepackage[colorlinks, linkcolor=black,  citecolor=black, urlcolor=black]{hyperref}
\usepackage{etoolbox}
\usepackage[ruled,linesnumbered]{algorithm2e}
\patchcmd\thebibliography
{\labelsep}
{\labelsep\itemsep=-1pt\relax}
{}
{\typeout{Couldn't patch the command}}

\title{\LARGE \bf Automatic hippocampal surface generation via 3D U-net and active shape modeling with hybrid particle swarm optimization}

\author{Pinyuan Zhong$^{1}$, Yue Zhang$^{1,2}$ and Xiaoying Tang$^{1,*}$
\thanks{*This study was supported by the National Natural Science Foundation of China (62071210), the Shenzhen Basic Research Program (JCYJ20190809120205578), the National Key R\&D Program of China (2017YFC0112404), and the High-level University Fund (G02236002).}
\thanks{*Correspondence at \tt\small \url{tangxy@sustech.edu.cn}}
\thanks{$^1$ Department of Electrical and Electronic Engineering, Southern University of Science and Technology, Shenzhen, China}%
\thanks{$^2$ Department of Electrical and Electronic Engineering, The University of Hong Kong, Hong Kong, China}%
}

\begin{document}

\maketitle
\thispagestyle{empty}
\pagestyle{empty}

\begin{abstract}
In this paper, we proposed and validated a fully automatic pipeline for hippocampal surface generation via 3D U-net coupled with active shape modeling (ASM). 
Principally, the proposed pipeline consisted of three steps. 
In the beginning, for each magnetic resonance image, a 3D U-net was employed to obtain the automatic hippocampus segmentation at each hemisphere. 
Secondly, ASM was performed on a group of pre-obtained template surfaces to generate mean shape and shape variation parameters through principal component analysis. 
Ultimately, hybrid particle swarm optimization was utilized to search for the optimal shape variation parameters that best match the segmentation. 
The hippocampal surface was then generated from the mean shape and the shape variation parameters. The proposed pipeline was observed to provide hippocampal surfaces at both hemispheres with high accuracy, correct anatomical topology, and sufficient smoothness.
\newline

\indent \textit{Clinical Relevance}— This work provides a useful tool for generating hippocampal surfaces which are important biomarkers for a variety of brain disorders.
\end{abstract}

\vspace{0.4em}
\section{INTRODUCTION}
\raggedbottom
The hippocampus has a close relationship with the memory function of the human brain.
Its volume reduction and morphological degeneration are tightly linked to many neurodegenerative diseases, such as Alzheimer’s disease (AD), Huntington’s disease, and other forms of dementia \cite{qiu2009regional,miller2015amygdalar,van2011shape,faria2016linking}. 
During the past decade, the hippocampal surface has become a very important object of interest in various morphometric analyses, considering the importance of localizing and quantifying region-specific morphometric abnormalities of the hippocampus in various memory-related brain disorders, especially at an early stage of the disease of interest \cite{zou2019regional}. 
For AD, a disease that links tightly to the hippocampus, the effectiveness of the hippocampal surface, in terms of revealing region-specific shape abnormalities and delivering discriminating shape features, has been identified in various previous studies \cite{tang2014shape}.

The hippocampal surfaces are typically created by directly applying mesh generation algorithms to the corresponding volumetric segmentations generated by, for example, deep learning based segmentation methods \cite{ronneberger2015u}. 
One of the most representative mesh generation approaches is the marching cubes algorithm, which has been widely used in medical image visualization and computer graphics \cite{qiu2008multi}.  
It transforms a volumetric segmentation’s boundary into a surface represented by a triangulated mesh that consists of vertices and faces.

A surface generated by the marching cubes algorithm tightly contours the corresponding volumetric segmentation. 
As such, the resulting surface is extremely sensitive to segmentation noises and may be severely affected by inaccurate segmentation. 
Such a characteristic may lead to incorrect anatomical topology such as holes or disconnected regions. 
What's more, the resulting surface is typically not smooth.
End-to-end approaches that directly generate hippocampal surfaces also exist, such as FSL-FIRST which employs a Bayesian shape model \cite{patenaude2011bayesian}. The surfaces generated by FSL-FIRST are smooth but not highly accurate.
Therefore, an approach that can generate surfaces with high accuracy, correct anatomical topology, and sufficient smoothness is needed.

In this paper, we propose a fully automatic pipeline for hippocampal surface generation. 
The volumetric segmentation of the hippocampus is first obtained utilizing a 3D U-net. 
The purpose of this step is to use a sophisticated neural network to obtain decent hippocampal segmentation results for guiding subsequent steps. 
Next, a large deformation diffeomorphic metric mapping (LDDMM) pipeline is applied to generate a group of template surfaces with sufficient smoothness \cite{tang2018fully}. 
Afterwards, active shape modeling (ASM) is performed to compute the mean shape and shape variation parameters of those template surfaces via principal component analysis (PCA) \cite{cootes1995active}. 
In the end, we search for the optimal shape variation parameters which best match the segmentation via a hybrid particle swarm optimization (PSO) algorithm with a Dice loss being its fitness function \cite{zhang2015comprehensive}. 
The mean shape and the deformation carried out according to the shape variation parameters are then combined to obtain the final surface.
The hippocampal surfaces generated by the proposed pipeline are shown to possess high accuracy, correct anatomical topology, and sufficient smoothness.

\vspace{0.4em}
\section{METHOD}
\begin{figure}[htbp]
	\centering
	\includegraphics[width=1\linewidth]{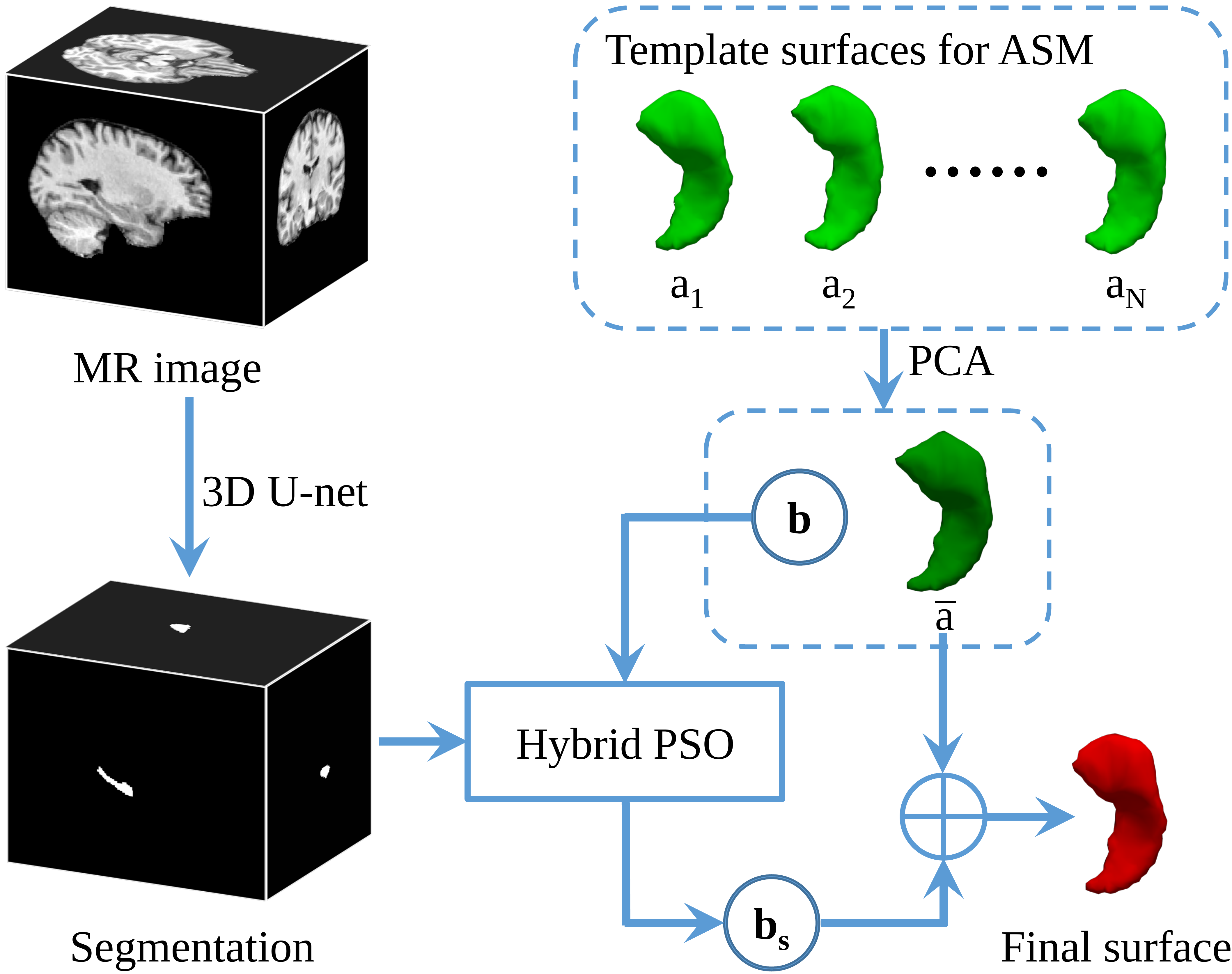}
	\caption{Flowchart of the proposed pipeline.}
	\label{fig:pipeline1}
\end{figure}

The flowchart of our proposed pipeline for hippocampal surface generation is illustrated in Fig. \ref{fig:pipeline1}. 

\subsection{Hippocampal segmentation}
For image preprocessing, all MR images are resampled to $190 \times 230 \times 180$ voxels of size $1 \times 1 \times 1$ mm$^3$. 
They are then skull-stripped, rigidly aligned to the MNI space, and input into a 3D U-net to obtain hippocampal segmentation results.
This 3D U-net is driven by nnU-net that can automatically adapt its architecture to the given image geometry \cite{isensee2018nnu}.
The network is trained with a combination of Dice loss and cross-entropy loss.
Please note the goal of the proposed pipeline is \textbf{automatic hippocampal surface generation}.
A utilization of the 3D U-net is only to obtain high-quality segmentation results to guide the subsequent surface generation process.
Theoretically, our employed nnU-net can be replaced with other segmentation approaches. However, considering its superior segmentation performance \cite{isensee2018nnu}, we select it to be our segmentation module. 

\vspace{0.5em}
\subsection{ASM}
After 3D U-net, a surface generation approach is applied to the segmentation results to generate template surfaces with sufficient smoothness, being prepared for the next shape model building procedure. 
Due to ASM's being an explicit shape model for constraining shape deformation, the resulting surfaces are relatively insensitive to noises and artifacts, providing consistent smoothness and correct anatomical topology as the template surfaces.

\subsubsection{Template surface generation}\label{g}
A LDDMM pipeline \cite{tang2018fully} is applied to the triangulated meshes (target surfaces) generated by the marching cubes algorithm from the 3D U-net’s segmentation results. 
During this process, an initial template surface with correct anatomical topology and a high degree of smoothness is adopted. 
Because LDDMM produces diffeomorphisms, the anatomical topology and smoothness of the initial template surface are inherited by the target surfaces. 
Also, diffeomorphisms derived from LDDMM are highly accurate, ensuring that the deformed template surfaces are very close to the target surfaces and thus of high fidelity. 
Last but not least, diffeomorphisms ensure equal numbers of vertices and one-to-one vertex correspondences across different target surfaces, being compatible with the subsequent ASM.

\subsubsection{PCA}
Each obtained template surface can be expressed as a shape vector with its coordinates of vertices aligned into a row vector.
Let $\textbf{a}_i$ be a vector describing the $n$ points of the $i^{th}$ template surface
\begin{equation}
\textbf{a}_i=(x_{i1},y_{i1},z_{i1},x_{i2},y_{i2},z_{i2},...,x_{in},y_{in},z_{in}).
\end{equation}
PCA is applied to all shape vectors to identify directions of maximum variation, and a small number of principal components are selected to reduce shape dimensionality, namely
\begin{equation}
\bar{\textbf{a}} = \frac{1}{{N}}\sum\limits_{i = 1}^N {\textbf{a}_i}
\end{equation}
\begin{equation}
\textbf{S} = \frac{1}{N}\sum\limits_{i = 1}^N {{{({\textbf{a}_i} - \bar{\textbf{a}})}^T}({\textbf{a}_i} - \bar{\textbf{a}})}
\end{equation}
where $\bar{\textbf{a}}$ is the mean shape and $\textbf{S}$ is the covariance matrix.
We then use $\textbf{Sv} = \lambda \textbf{v}$ to compute all possible solutions of the eigenvalue $\lambda$ and the eigenvector $\textbf{v}$.
Generally, most variation can be explained by a small number of modes, $t$. 
The first $t$ eigenvectors compose a transfer matrix $\textbf{P}=(\textbf{v}_1,\textbf{v}_2,...,\textbf{v}_t)$. 
And any shape vector can then be approximated by using the mean shape and a weighted sum of the first $t$ eigenvectors
\begin{equation}
\textbf{a}=\bar{\textbf{a}}+\textbf{Pb},
\end{equation}
where $\textbf{b}=(b_1,b_2,...,b_t)^T$ is a vector of shape parameters and it controls the variation of the shape.
We can thus obtain a new surface by varying the parameters of $\textbf{b}$.
Besides these $t$ shape parameters, we use seven extra parameters to model an additional transformation with three parameters for translation, three parameters for rotation, and one parameter for isotropic scaling.

\vspace{0.6em}
\subsection{Hybrid PSO searching}
A hybrid PSO algorithm, composed of global best and local best, is applied to search for the optimal $t+7$ parameters using a Dice loss with the segmentation results from 3D U-net as its fitness function.
These parameters here are considered as the position of one particle.
At the first stage, the particle swarm is initialized with a population of random solutions. 
During each iteration the particle searches for optima by updating its velocity and position with the following formulas
\begin{equation}
\begin{split}
\textbf{V}_i^{(k+1)}=w\textbf{V}_i^{(k)}+c_1r_1(\textbf{P}_i^{(k)}-\textbf{X}_i^{(k)})+c_2\\
r_2[\alpha(\textbf{P}_g^{(k)}-\textbf{X}_i^{(k)})+(1-\alpha)(\textbf{P}_l^{(k)}-\textbf{X}_i^{(k)})],
\end{split}
\end{equation}
\begin{equation}
\textbf{X}_i^{(k+1)}=\textbf{X}_i^{(k)}+\textbf{V}_i^{(k+1)}
\end{equation}
where $\textbf{V}_i^{(k)}$ and $\textbf{X}_i^{(k)}$ respectively represent the velocity and position of the $i^{th}$ particle after the $k^{th}$ iteration; $\textbf{P}_i^{(k)}$, $\textbf{P}_g^{(k)}$ and $\textbf{P}_l^{(k)}$ respectively denote the best position of the $i^{th}$ particle, the best position of the global particles, and that of the local particles within the neighborhood after the $k^{th}$ iteration; $w, \alpha \in [0, 1]$ respectively denote an inertia weight and a weight index to balance the effect between global best and local best; $r_1, r_2 \sim U(0, 1)$ are two random numbers both following uniform distributions; $c_1$, $c_2$ are acceleration constants, which control how far a particle moves in one iteration. 
Compared with the standard PSO, the hybrid PSO can well solve the issue of converging to a local optimal solution.  
After the searching process, the optimized parameters $\textbf{b}_s$ are combined with the mean shape $\bar{\textbf{a}}$ to generate the final surface.

\vspace{0.5em}
\section{EXPERIMENTS AND RESULTS}
\subsection{Dataset and evaluation criteria}
The proposed pipeline is evaluated on a dataset obtained from the PREDICT-HD study (\href{https://www.predict-hd.net/}{https://www.predict-hd.net/}) consisting of 16 T1-weighted images with manual annotations of the bilateral hippocampi. At the stage of automatic segmentation, we employ a four-fold cross-validation strategy, wherein each fold contains 12 images for training and four images for testing. After that, a LDDMM pipeline is applied to all segmentation results followed by ASM, suggesting that each individual image performs its hippocampal surface generation using a shape model built from all 16 images, without any use of the ground truth labeling.

\begin{figure*}[hbp]
	\centering
	\includegraphics[width=1\linewidth]{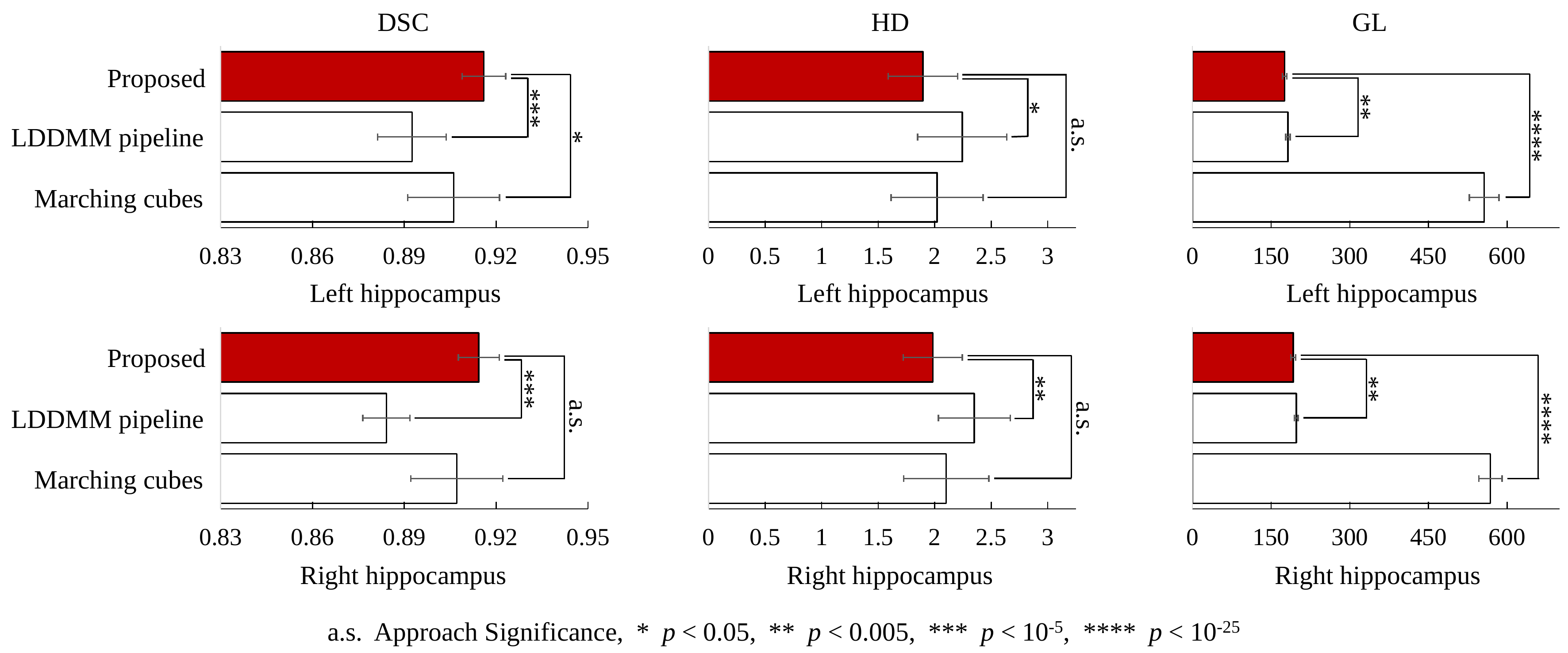}
	\caption{Quantitative comparisons of the marching cubes algorithm, the LDDMM pipeline, and the proposed pipeline in hippocampal surface generation.}
	\label{fig:pipeline2}
\end{figure*}

To quantify the accuracy of the proposed pipeline, we use the Dice similarity coefficient (DSC) \cite{dice1945measures} and the Hausdorff distance (HD) \cite{huttenlocher1993comparing} between the voxelization of the obtained surface and the corresponding manual delineation. The geometric laplacian (GL) is used to quantify the smoothness of a surface
\begin{equation}
GL(v) = v - \frac{{\sum\nolimits_{i \in n(v)} {l_i^{ - 1}{v_i}} }}{{\sum\nolimits_{i \in n(v)} {l_i^{ - 1}} }},
\end{equation}
where vertices $v_i$ represent the direct neighbors of $v$, $n(v)$ is the index set of these neighbors, and $l_i$ is the Euclidean distance from $v$ to $v_i$. $GL(v)$ denotes the roughness of the surface around $v$, and a lower value reflects a smoother surface. The sum of the norm of all vertex-wise GL vectors is computed as the GL of the entire surface: $GL = \sum\limits_v {{{\left\| {GL(v)} \right\|}_2}}$.

\subsection{Results and discussion}
To demonstrate the superiority of our proposed pipeline, we compare it with the marching cubes algorithm, a LDDMM pipeline, and a state-of-the-art method FSL-FIRST \cite{patenaude2011bayesian}. Please note both the marching cubes algorithm and the LDDMM pipeline create surfaces from our 3D U-net segmentations, whereas FSL-FIRST is a completely standalone pipeline that delivers both segmentations and surfaces on its own. Quantitative comparisons of the first two surface generation methods and our proposed pipeline in terms of DSC, HD, and GL are shown in Fig. \ref{fig:pipeline2}. 
Overall, the proposed pipeline outperforms both the marching cubes algorithm and the LDDMM pipeline in terms of all three metrics. Specifically, for the left hippocampus, the proposed pipeline is statistically significantly better than both the marching cubes algorithm and the LDDMM pipeline in terms of accuracy (DSC) and smoothness (GL), and it is statistically significantly better than the LDDMM pipeline in terms of boundary distance (HD) while slightly better than the marching cubes algorithm in that respect. For the right hippocampus, the proposed pipeline is statistically significantly better than the LDDMM pipeline in terms of accuracy (DSC), boundary distance (HD) and smoothness (GL), and it is statistically significantly better than the marching cubes algorithm in terms of smoothness (GL) while slightly better in terms of accuracy (DSC) and boundary distance (HD). Quantitative comparisons between FSL-FIRST and our proposed pipeline are listed in Table \ref{table1}. Evidently, the proposed pipeline performs much better than FSL-FIRST in terms of accuracy (DSC) and boundary distance (HD) and is similar in terms of smoothness (GL). Collectively, the proposed pipeline provides a very good balance between accuracy and smoothness.

\begin{figure}[htbp]
	\centering
	\includegraphics[width=1\linewidth]{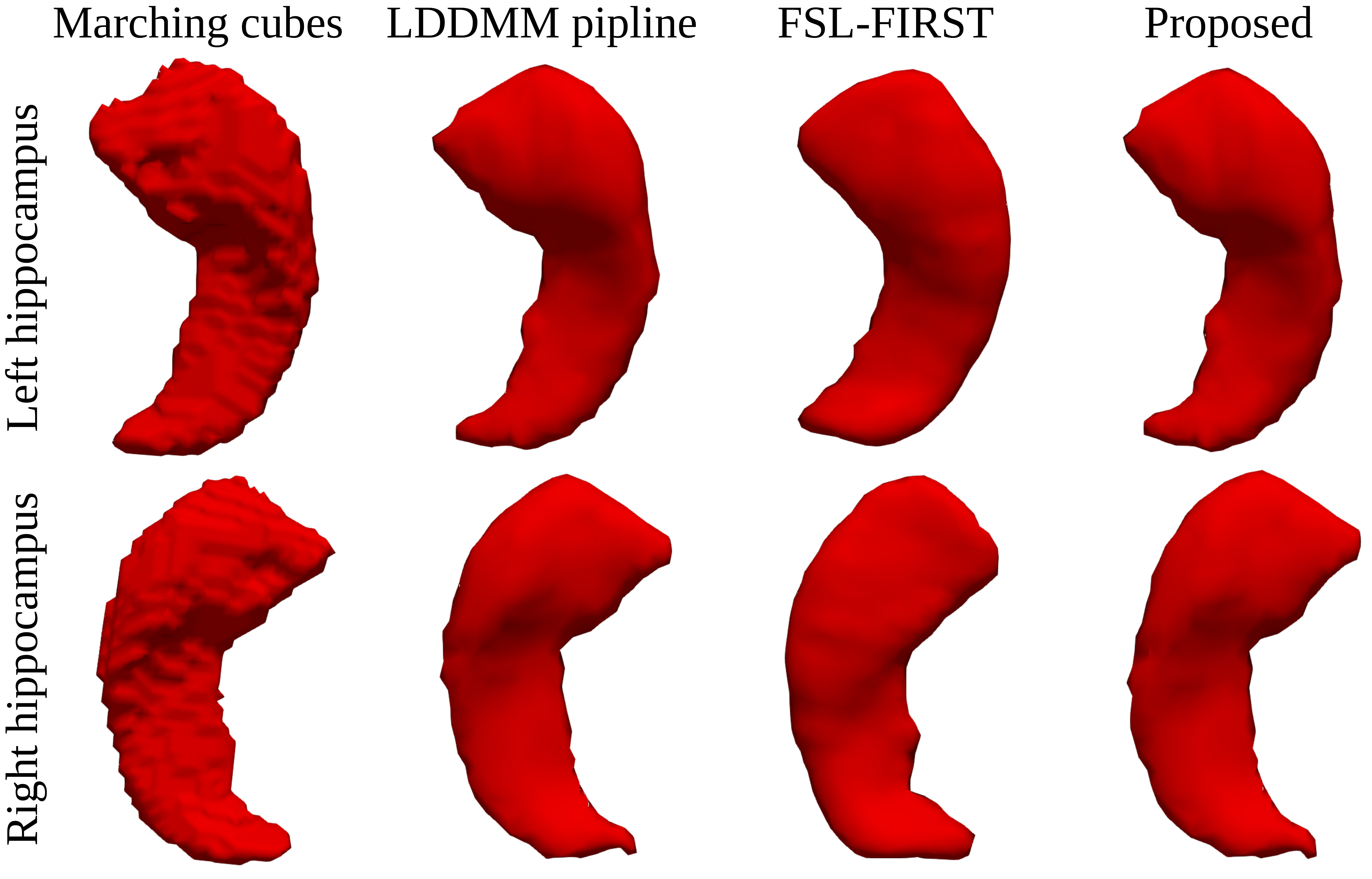}
	\caption{Representative hippocampal surfaces generated from four methods.}
	\label{fig:pipeline3}
\end{figure}

A visual comparison of representative surfaces generated by these four methods is demonstrated in Fig. \ref{fig:pipeline3}. For surfaces generated by the marching cubes algorithm, they have high fidelity to the volumetric segmentations but are typically topology incorrect and very rough. For surfaces generated by FSL-FIRST, they typically have sufficient smoothness but relatively low accuracy. As for the LDDMM pipeline, the generated surfaces possess high accuracy and sufficient smoothness. However, it is significantly outperformed by our proposed pipeline in both aspects. Through such comprehensive comparisons, the proposed pipeline performs supremely in generating surfaces with high accuracy, correct anatomical topology, and sufficient smoothness. To further validate the effectiveness and robustness of the proposed pipeline, testing it on clinical datasets with more structures of interest will be needed. We anticipate that as one of our future endeavors.

\begin{table*}[!ht]
	\centering
	\footnotesize    
	\renewcommand{\arraystretch}{1.2}   
	\setlength{\abovecaptionskip}{0ex}%
	\setlength{\belowcaptionskip}{2pt}%
	\caption{The mean and standard deviation values of DSC, HD, and GL obtained from FSL-FIRST and the proposed pipeline. Bold indicates the better result in terms of the mean value. Keys: L-Hippo -- left hippocampus, R-Hippo -- right hippocampus.}
	\medskip
	\label{table1}
	\begin{tabular}{lcccccc}  \hline
		& \multicolumn{2}{c}{DSC} & \multicolumn{2}{c}{HD} & \multicolumn{2}{c}{GL}\\ \cmidrule(r){2-3} \cmidrule(r){4-5} \cmidrule(r){6-7}
		& L-Hippo  & R-Hippo & L-Hippo  & R-Hippo & L-Hippo  & R-Hippo\\
		\hline
		FSL-FIRST \cite{patenaude2011bayesian}   & 0.817 $\pm$ 0.018	& 0.831 $\pm$ 0.019 & 3.750 $\pm$ 0.829	& 3.335 $\pm$ 0.938 & \textbf{172.725 $\pm$ 7.591}	& 194.880 $\pm$ 9.232\\
		Proposed & \textbf{0.916 $\pm$ 0.007}	& \textbf{0.914 $\pm$ 0.007} & \textbf{1.895 $\pm$ 0.307}	& \textbf{1.983 $\pm$ 0.261} & 176.415 $\pm$ 4.109	& \textbf{192.206 $\pm$ 4.250}\\
		\emph{p}-value & 1.505E-18 & 4.000E-16 & 4.570E-09 & 8.050E-06 & 0.162 & 0.326\\ 
		\hline		
	\end{tabular}
\end{table*}

\section{CONCLUSION}
In this paper, we proposed and validated a fully automatic pipeline for hippocampal surface generation utilizing 3D U-net coupled with ASM. A LDDMM pipeline is used to generate multiple template surfaces and a hybrid PSO algorithm is applied to search for the optimized shape parameters in ASM. Our experimental results demonstrate the superiority of the proposed pipeline in generating hippocampal surfaces with high accuracy, correct anatomical topology, and sufficient smoothness.

\LARGE
\bibliographystyle{IEEEbib}
\bibliography{strings,refs}

\begin{thebibliography}{10}

\bibitem{qiu2009regional}
Anqi Qiu, Christine Fennema-Notestine, Anders~M Dale, Michael~I Miller,
  Alzheimer's Disease~Neuroimaging Initiative, et~al.,
\newblock ``Regional shape abnormalities in mild cognitive impairment and
  alzheimer's disease,''
\newblock {\em Neuroimage}, vol. 45, no. 3, pp. 656--661, 2009.

\bibitem{miller2015amygdalar}
Michael~I Miller, Laurent Younes, J~Tilak Ratnanather, Timothy Brown, Huong
  Trinh, David~S Lee, Daniel Tward, Pamela~B Mahon, Susumu Mori, Marilyn
  Albert, et~al.,
\newblock ``Amygdalar atrophy in symptomatic alzheimer's disease based on
  diffeomorphometry: the biocard cohort,''
\newblock {\em Neurobiology of aging}, vol. 36, pp. S3--S10, 2015.

\bibitem{van2011shape}
Simon~JA van~den Bogaard, Eve~M Dumas, Luca Ferrarini, Julien Milles, et~al.,
\newblock ``Shape analysis of subcortical nuclei in huntington's disease,
  global versus local atrophy—results from the track-hd study,''
\newblock {\em Journal of the neurological sciences}, vol. 307, no. 1-2, pp.
  60--68, 2011.

\bibitem{faria2016linking}
Andreia~V Faria, J~Tilak Ratnanather, Daniel~J Tward, et~al.,
\newblock ``Linking white matter and deep gray matter alterations in
  premanifest huntington disease,''
\newblock {\em Neuroimage: clinical}, vol. 11, pp. 450--460, 2016.

\bibitem{zou2019regional}
Lin Zou, Yukun Song, Xiangxue Zhou, Jianping Chu, and Xiaoying Tang,
\newblock ``Regional morphometric abnormalities and clinical relevance in
  wilson's disease,''
\newblock {\em Movement disorders}, vol. 34, no. 4, pp. 545--554, 2019.

\bibitem{tang2014shape}
Xiaoying Tang, Dominic Holland, Anders~M Dale, Laurent Younes, Michael~I
  Miller, and Alzheimer's Disease~Neuroimaging Initiative,
\newblock ``Shape abnormalities of subcortical and ventricular structures in
  mild cognitive impairment and alzheimer's disease: detecting, quantifying,
  and predicting,''
\newblock {\em Human brain mapping}, vol. 35, no. 8, pp. 3701--3725, 2014.

\bibitem{ronneberger2015u}
Olaf Ronneberger, Philipp Fischer, and Thomas Brox,
\newblock ``U-net: Convolutional networks for biomedical image segmentation,''
\newblock in {\em International conference on medical image computing and
  computer-assisted intervention}. Springer, 2015, pp. 234--241.

\bibitem{qiu2008multi}
Anqi Qiu and Michael~I Miller,
\newblock ``Multi-structure network shape analysis via normal surface momentum
  maps,''
\newblock {\em Neuroimage}, vol. 42, no. 4, pp. 1430--1438, 2008.

\bibitem{patenaude2011bayesian}
Brian Patenaude, Stephen~M Smith, David~N Kennedy, and Mark Jenkinson,
\newblock ``A bayesian model of shape and appearance for subcortical brain
  segmentation,''
\newblock {\em Neuroimage}, vol. 56, no. 3, pp. 907--922, 2011.

\bibitem{tang2018fully}
Xiaoying Tang, Yuan Luo, Zhibin Chen, et~al.,
\newblock ``A fully-automated subcortical and ventricular shape generation
  pipeline preserving smoothness and anatomical topology,''
\newblock {\em Frontiers in neuroscience}, vol. 12, pp. 321, 2018.

\bibitem{cootes1995active}
Timothy~F Cootes, Christopher~J Taylor, David~H Cooper, and Jim Graham,
\newblock ``Active shape models-their training and application,''
\newblock {\em Computer vision and image understanding}, vol. 61, no. 1, pp.
  38--59, 1995.

\bibitem{zhang2015comprehensive}
Yudong Zhang, Shuihua Wang, and Genlin Ji,
\newblock ``A comprehensive survey on particle swarm optimization algorithm and
  its applications,''
\newblock {\em Mathematical problems in engineering}, vol. 2015, 2015.

\bibitem{isensee2018nnu}
Fabian Isensee, Jens Petersen, Andre Klein, David Zimmerer, Paul~F Jaeger,
  Simon Kohl, Jakob Wasserthal, Gregor Koehler, Tobias Norajitra, Sebastian
  Wirkert, et~al.,
\newblock ``nnu-net: Self-adapting framework for u-net-based medical image
  segmentation,''
\newblock {\em arXiv preprint arXiv:1809.10486}, 2018.

\bibitem{dice1945measures}
Lee~R Dice,
\newblock ``Measures of the amount of ecologic association between species,''
\newblock {\em Ecology}, vol. 26, no. 3, pp. 297--302, 1945.

\bibitem{huttenlocher1993comparing}
Daniel~P Huttenlocher, Gregory~A. Klanderman, and William~J Rucklidge,
\newblock ``Comparing images using the hausdorff distance,''
\newblock {\em IEEE transactions on pattern analysis and machine intelligence},
  vol. 15, no. 9, pp. 850--863, 1993.

\end{thebibliography}

\end{document}